\documentstyle[aclap]{article}

\title{\vspace{-0.5in}An Efficient Generation Algorithm for Lexicalist MT}
\author{Victor Pozna\'{n}ski, John L. Beaven \& Pete Whitelock \thanks{We wish
to thank our colleagues Kerima Benkerimi, David Elworthy, Peter Gibbins, Ian
Johnson, Andrew Kay and Antonio Sanfilippo at SLE, and our anonymous
reviewers for useful feedback
and discussions on the research reported here and on earlier drafts of this
paper. }\\
SHARP Laboratories of Europe Ltd. \\
Oxford Science Park, Oxford OX4 4GA \\
United Kingdom \\
\{vp,jlb,pete\}@sharp.co.uk}

\begin{document}
\maketitle
\vspace{-0.5in}

\begin{abstract}

The lexicalist approach to Machine Translation offers significant
advantages in the development of linguistic descriptions.  However, the
Shake-and-Bake generation algorithm of \cite{Whitelock:S&B} is
NP-complete.  We present a polynomial time algorithm for lexicalist MT
generation provided that sufficient information can be transferred to
ensure more determinism.

\end{abstract}

\section{Introduction}

Lexicalist approaches to MT, particularly those incorporating the
technique of {\em Shake-and-Bake\/} generation
\cite{Beaven:Lexicalist,Beaven:S&B,Whitelock:1994}, combine the linguistic
advantages of transfer \cite{Arnold:Relaxed,Allegranza:Eurotra} and
interlingual \cite{Nirenburg:MTKB,Dorr:MT} approaches.  Unfortunately,
the generation algorithms described to date have been intractable.  In
this paper, we describe an alternative generation component which has
polynomial time complexity.

Shake-and-Bake translation assumes a source grammar, a target grammar and a
bilingual dictionary which relates translationally equivalent sets of lexical
signs, carrying across the semantic dependencies established by the source
language analysis stage into the target language generation stage.

The translation process consists of three phases:
\begin{enumerate}
\item A {\it parsing phase}, which outputs a multiset,
or {\it bag}, of source language signs instantiated with sufficiently rich
linguistic information established by the parse to ensure adequate
translations.
\item A {\it lexical-semantic transfer phase} which employs the bilingual
dictionary to map the bag of instantiated source signs onto a bag of
target language signs.
\item A  {\it generation phase} which imposes an order on the bag of
target signs which is guaranteed grammatical according
to the monolingual target grammar. This ordering must respect the
linguistic constraints which have been transferred into the target signs.
\end{enumerate}

The {\it Shake-and-Bake} generation algorithm of \cite{Whitelock:S&B} combines
target language signs using the technique known as {\it generate-and-test}. In
effect, an arbitrary permutation of signs is input to a shift-reduce parser
which tests them for grammatical well-formedness. If they are well-formed, the
system halts indicating success. If not, another permutation is tried and the
process repeated.  The complexity of this algorithm is $O(n!)$ because all
permutations ($n! $ for an input of size $n$) may have to be explored to find
the correct answer, and indeed {\em must} be explored in order to verify that
there is no answer.

Proponents of the Shake-and-Bake approach have employed various techniques to
improve generation efficiency. For example, \cite{Beaven:Lexicalist} employs a
chart to avoid recalculating the same combinations of signs more than once
during testing, and \cite{Popowich:Efficiency} proposes a more general
technique for storing which rule applications have been attempted;
\cite{Brew:Cat} avoids certain pathological cases by employing
global constraints on the solution space; researchers such as
\cite{Brown:Statistical} and \cite{Chen:BagGen} provide a system for bag
generation that is heuristically guided by probabilities. However, none of
these approaches is guaranteed to avoid protracted search times if an exact
answer is required, because bag generation is NP-complete \cite{Brew:Cat}.

Our novel generation algorithm has polynomial complexity ($O(n^4)$). The
reduction in theoretical complexity is achieved by placing constraints on the
power of the target grammar when operating on instantiated signs, and by using
a more restrictive data structure than a bag, which we call a {\it target
language normalised commutative bracketing (TNCB)}. A TNCB records dominance
information from derivations and is amenable to incremental updates. This
allows us to employ a greedy algorithm to refine the structure progressively
until either a target constituent is found and generation has succeeded or no
more changes can be made and generation has failed.

In the following sections, we will sketch the basic algorithm, consider how to
provide it with an initial guess, and provide an informal proof of its
efficiency.

\section{A Greedy Incremental Generation Algorithm}

We begin by describing the fundamentals of a greedy incremental generation
algorithm. The crucial data structure that it employs is the {\em TNCB}.  We
give some definitions, state some key assumptions about
suitable TNCBs for generation, and then describe the algorithm itself.

\subsection{TNCBs}

We assume a sign-based grammar with binary rules, each of which may be used to
{\it combine} two signs by unifying them with the daughter categories and
returning the mother. Combination is the commutative equivalent of rule
application; the linear ordering of the daughters that leads to
successful rule application determines the orthography of the mother.

Whitelock's Shake-and-Bake generation algorithm attempts to arrange the bag of
target signs until a grammatical ordering (an ordering which allows all of the
signs to combine to yield a single sign) is found. However, the target {\it
derivation} information itself is not used to assist the algorithm. Even in
\cite{Beaven:Lexicalist}, the derivation information is used simply to cache
previous results to avoid exact recomputation at a later stage, not to improve
on previous guesses. The reason why we believe such improvement is possible is
that, given adequate information from the previous stages, two
target signs cannot combine by accident; they must do so because the
underlying semantics within the signs licenses it.

If the linguistic data that two signs contain allows them to
combine, it is because they are providing a semantics which might later become
more specified. For example, consider the bag of signs that have been derived
through the Shake-and-Bake process which represent the phrase: \\
\\
(1) The big brown dog \\
\\
Now, since the determiner and adjectives all modify the same noun, most
grammars will allow us to construct the phrases: \\
\\
(2) The dog\\
(3) The big dog \\
(4) The brown dog\\
\\
as well as the `correct' one.  Generation will fail if all
signs in the bag are not eventually incorporated in the final result,
but in the na\"{\i}ve algorithm, the intervening computation may be
intractable.

In the algorithm presented here, we start from observation that the phrases
(2) to (4) are not incorrect semantically; they are
simply under-specifications of (1). We take advantage of this by recording
the constituents that have combined within the TNCB, which is designed to
allow further constituents to be incorporated with minimal recomputation.

A TNCB is composed of a sign, and a history of how it was derived from its
children. The structure is essentially a binary derivation tree whose
children are unordered.  Concretely, it is either NIL, or a triple:

\begin{eqnarray*}
\mbox{TNCB} & = &  \mbox{NIL}\, | \, \mbox{Value} \times \mbox{TNCB} \times
\mbox{TNCB} \\
\mbox{Value} & =  & \mbox{Sign}\, | \\
	& & \mbox{INCONSISTENT}\, | \\
	& & \mbox{UNDETERMINED}
\end{eqnarray*}

The second and third items of the TNCB triple are the {\em child TNCBs}.
The {\it value} of a TNCB is the sign that is formed from the combination of
its children, or {\it INCONSISTENT}, representing the fact that they
cannot grammatically combine, or {\it UNDETERMINED},  i.e. it has not
yet been established whether the signs combine.

\begin{figure}[htbp]
\begin{center}
{\tt    \setlength{\unitlength}{0.60pt}
\begin{picture}(291,81)
\thinlines    \put(255,66){\circle*{8}}
              \put(169,47){\circle*{8}}
              \put(107,67){\circle*{8}}
              \put(47,43){\circle*{8}}
              \put(245,46){\circle*{8}}
              \put(118,45){\circle*{8}}
              \put(36,67){\circle*{8}}
              \put(179,67){\circle*{8}}
              \put(212,34){=}
              \put(138,34){=}
              \put(66,34){=}
              \put(10,12){S}
              \put(55,12){O}
              \put(32,12){V}
              \put(274,12){S}
              \put(230,12){O}
              \put(255,12){V}
              \put(233,23){\line(1,2){22}}
              \put(254,68){\line(1,-2){22}}
              \put(245,46){\line(1,-2){11}}
              \put(14,23){\line(1,2){22}}
              \put(35,68){\line(1,-2){22}}
              \put(35,23){\line(1,2){11}}
              \put(169,47){\line(1,-2){11}}
              \put(178,69){\line(1,-2){22}}
              \put(157,24){\line(1,2){22}}
              \put(81,12){S}
              \put(103,12){O}
              \put(126,12){V}
              \put(106,23){\line(1,2){11}}
              \put(106,69){\line(1,-2){22}}
              \put(85,24){\line(1,2){22}}
              \put(154,12){V}
              \put(178,12){O}
              \put(198,12){S}
\end{picture}}
\end{center}
\caption{Equivalent TNCBs} \label{equivalences}
\end{figure}
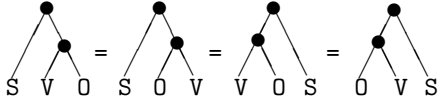

Undetermined TNCBs are commutative, e.g. they do not distinguish between
the structures shown in Figure \ref{equivalences}.

In section \ref{initialisation} we will see that this property is important
when starting up the generation process.

Let us introduce some terminology.

A TNCB is
\begin{itemize}
\item {\it well-formed} iff its value is a sign,
\item {\it ill-formed} iff its value is INCONSISTENT,
\item {\it undetermined} (and its value is UNDETERMINED) iff it has not
been demonstrated whether it is well-formed or ill-formed.
\item {\it maximal} iff it is well-formed and its parent (if it has one)
is ill-formed. In other words, a maximal TNCB is a largest well-formed
component of a TNCB.

Since TNCBs are tree-like structures, if a TNCB is
undetermined or ill-formed then so are all of its ancestors (the TNCBs that
contain it).
\end{itemize}

We define five operations on a TNCB. The first three are used to define the
fourth transformation ({\em move\/}) which improves ill-formed TNCBs. The
fifth is used to establish the well-formedness of undetermined nodes.
In the diagrams, we use a cross to represent ill-formed nodes and a black
circle to represent undetermined ones.

\begin{itemize} \item[] {\bf Deletion:} A maximal TNCB can be deleted
from its current position.  The structure above it must be adjusted in
order to maintain binary branching.  In figure \ref{deletion}, we see
that when node 4 is deleted, so is its parent node 3.  The new node 6,
representing the combination of 2 and 5, is marked undetermined.

\begin{figure}[htbp]
\begin{center}
{\tt    \setlength{\unitlength}{0.60pt}
\begin{picture}(324,124)
\thicklines   \put(248,106){6}
\thinlines    \put(252,98){\circle*{10}}
\thicklines   \put(47,10){\dashbox{5}(35,25){}}
\thinlines    \put(307,19){5}
              \put(193,19){2}
              \put(124,19){5}
              \put(61,19){4}
              \put(94,74){3}
              \put(10,19){2}
              \put(67,103){1}
              \put(252,99){\line(-4,-5){54}}
              \put(252,99){\line(5,-6){54}}
\thicklines   \put(129,70){\vector(1,0){54}}
\thinlines    \put(96,68){\line(-4,-5){28}}
              \put(69,100){\line(5,-6){54}}
              \put(69,100){\line(-4,-5){54}}
              \put(54,92){\begin{picture}(30,15)
\thicklines       \put(29,13){\line(-5,-2){28}}
                  \put(1,13){\line(5,-2){28}}
                  \end{picture}}
              \put(81,60){\begin{picture}(30,15)
\thicklines       \put(29,13){\line(-5,-2){28}}
                  \put(1,13){\line(5,-2){28}}
                  \end{picture}}
\end{picture}}
\end{center}
\caption{4 is deleted, raising 5} \label{deletion}
\end{figure}
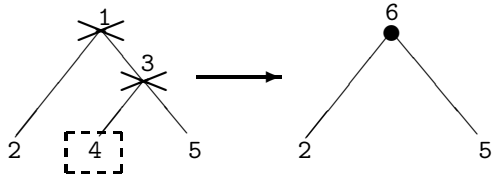

\item[] {\bf Conjunction:} A maximal TNCB can be conjoined with another
maximal TNCB if they may be combined by rule.  In figure
\ref{conjunction}, it can be seen how the maximal TNCB composed of nodes
1, 2, and 3 is conjoined with the maximal TNCB composed of nodes 4, 5
and 6 giving the TNCB made up of nodes 1 to 7.  The new node, 7, is
well-formed.

\begin{figure}[htbp]
\begin{center}
{\tt    \setlength{\unitlength}{0.60pt}
\begin{picture}(314,114)
\thinlines    \put(44,89){\line(1,-2){32}}
              \put(44,89){\line(-1,-2){32}}
              \put(150,88){\line(-1,-2){32}}
              \put(150,88){\line(1,-2){32}}
              \put(85,45){\begin{picture}(23,21)
\thicklines       \put(11,21){\line(0,-1){21}}
                  \put(0,11){\line(1,0){23}}
                  \end{picture}}
\thicklines   \put(183,56){\vector(1,0){41}}
\thinlines    \put(267,91){\line(1,-2){32}}
              \put(267,91){\line(-1,-2){32}}
              \put(251,55){\line(1,-2){14}}
              \put(285,55){\line(-1,-2){14}}
              \put(42,94){1}
              \put(10,12){2}
              \put(74,12){3}
              \put(146,91){4}
              \put(114,12){5}
              \put(178,12){6}
              \put(265,96){7}
              \put(243,57){1}
              \put(233,12){2}
              \put(258,12){3}
              \put(284,57){4}
              \put(268,12){5}
              \put(297,12){6}
\end{picture}}
\end{center}
\caption{1 is conjoined with 4 giving 7}
\label{conjunction}
\end{figure}
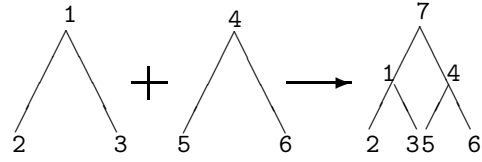

\item[] {\bf Adjunction:} A maximal TNCB can be inserted inside a
maximal TNCB, i.e.  conjoined with a non-maximal TNCB, where the
combination is licensed by rule.  In figure \ref{adjunction}, the TNCB
composed of nodes 1, 2, and 3 is inserted inside the TNCB composed of
nodes 4, 5 and 6.  All nodes (only 8 in figure \ref{adjunction})
which dominate the node corresponding to the new combination (node 7) must be
marked undetermined --- such nodes are said to be disrupted.

\begin{figure}[htbp]
\begin{center}
{\tt    \setlength{\unitlength}{0.50pt}
\begin{picture}(408,174)
\thinlines    \put(321,145){\circle*{12}}
              \put(50,118){\line(-1,-2){29}}
              \put(50,118){\line(1,-2){29}}
              \put(185,118){\line(1,-2){29}}
              \put(185,118){\line(-1,-2){29}}
              \put(320,145){\line(1,-2){59}}
              \put(320,145){\line(-1,-2){59}}
              \put(350,85){\line(-1,-2){28}}
              \put(337,57){\line(1,-2){14}}
\thicklines   \put(111,95){\line(0,-1){21}}
              \put(99,85){\line(1,0){24}}
              \put(227,85){\vector(1,0){44}}
              \put(217,51){\circle{22}}
              \put(47,122){1}
              \put(10,45){2}
              \put(80,45){3}
              \put(182,122){4}
              \put(146,45){5}
              \put(214,45){6}
              \put(355,94){7}
              \put(327,57){1}
              \put(317,12){2}
              \put(350,12){3}
              \put(256,12){5}
              \put(380,12){6}
              \put(318,156){8}
\end{picture}}
\end{center}
\caption{1 is adjoined next to 6 inside 4}
\label{adjunction}
\end{figure}
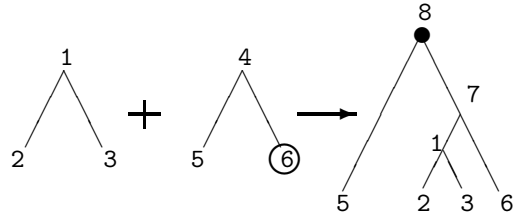

\item[] {\bf Movement:} This is a combination of a deletion with a subsequent
conjunction or adjunction. In figure \ref{movement}, we illustrate a move via
conjunction.  In the left-hand figure, we assume we wish to move the maximal
TNCB 4 next to the maximal TNCB 7. This first involves deleting TNCB 4
(noting it), and raising node 3 to replace node 2. We then introduce node
8 above node 7, and make both nodes 7 and 4 its children.  Note that
during deletion, we remove a surplus node (node 2 in this case) and during
conjunction or adjunction we introduce a new one (node 8 in this case) thus
maintaining the same number of nodes in the tree.

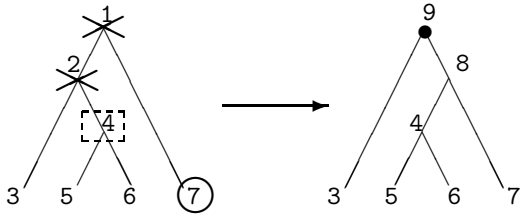
\begin{figure}[htbp]
\begin{center}
{\tt    \setlength{\unitlength}{0.45pt}
\begin{picture}(470,189)
\thicklines   \put(372,171){9}
\thinlines    \put(374,161){\circle*{12}}
              \put(104,164){\line(-1,-2){67}}
              \put(104,164){\line(1,-2){67}}
              \put(82,122){\line(1,-2){45}}
              \put(104,76){\line(-1,-2){22}}
              \put(373,164){\line(1,-2){67}}
              \put(374,164){\line(-1,-2){67}}
              \put(394,122){\line(-1,-2){45}}
              \put(372,76){\line(1,-2){22}}
              \put(86,69){\dashbox{5}(35,25){}}
\thicklines   \put(181,24){\circle{28}}
              \put(101,169){1}
              \put(73,127){2}
              \put(22,16){3}
              \put(102,78){4}
              \put(67,16){5}
              \put(120,16){6}
              \put(174,16){7}
              \put(400,128){8}
              \put(204,99){\vector(1,0){89}}
              \put(86,155){\begin{picture}(36,21)
\thicklines       \put(1,2){\line(2,1){33}}
                  \put(1,19){\line(2,-1){34}}
                  \end{picture}}
              \put(64,110){\begin{picture}(36,21)
\thicklines       \put(1,2){\line(2,1){33}}
                  \put(1,19){\line(2,-1){34}}
                  \end{picture}}
\thicklines   \put(292,16){3}
              \put(361,78){4}
              \put(340,16){5}
              \put(393,16){6}
              \put(443,16){7}
\end{picture}}
\end{center}
\caption{A conjoining move from 4 to 7}
\label{movement}
\end{figure}

\item[]{\bf Evaluation:} After a movement, the TNCB is undetermined as
demonstrated in figure \ref{movement}. The signs of the affected parts must be
recalculated by combining the recursively evaluated child TNCBs.
\end{itemize}

\subsection{Suitable Grammars}

The Shake-and-Bake system of \cite{Whitelock:S&B} employs a bag generation
algorithm because it is assumed that the input to the generator is no more
than a collection of instantiated signs. Full-scale bag
generation is not necessary because sufficient information can be transferred
from the source language to severely constrain the subsequent search during
generation.

The two properties required of TNCBs (and hence the target grammars with
instantiated lexical signs) are:

\begin{enumerate} \item {\bf Precedence Monotonicity.} The order of the
orthographies of two combining signs in the orthography of the result must be
determinate --- it must not depend on any subsequent combination that the
result may undergo. This constraint says that if one constituent
fails to combine with another, no permutation of the elements
making up either would render the combination possible.  This
allows bottom-up evaluation to occur in linear time. In practice, this
restriction requires that sufficiently rich information be transferred from
the previous translation stages to ensure that sign combination is
deterministic.

\item {\bf Dominance Monotonicity.}
If a maximal TNCB is adjoined at the highest possible place inside another
TNCB, the result will be well-formed after it is re-evaluated.  Adjunction is
only attempted if conjunction fails (in fact conjunction is merely a special
case of adjunction in which no nodes are disrupted); an adjunction which
disrupts $i$ nodes is attempted before one which disrupts $i+1$
nodes. Dominance monotonicity merely requires all nodes that are disrupted
under this top-down control regime to be well-formed when re-evaluated.  We
will see that this will ensure the termination of the generation algorithm
within $n-1$ steps, where $n$ is the number of lexical signs input to the
process.

\end{enumerate}

We are currently investigating the mathematical characterisation of grammars
and instantiated signs that obey these constraints.  So far, we have not
found these restrictions particularly problematic.

\subsection{The Generation Algorithm}

\begin{figure}[htbp]
\begin{center}
{\tt    \setlength{\unitlength}{0.60pt}
\begin{picture}(282,140)
\thinlines    \put(10,13){PAST}
              \put(59,13){dog}
              \put(103,13){bark}
              \put(153,13){the}
              \put(197,13){brown}
              \put(251,13){big}
              \put(22,22){\line(6,5){120}}
              \put(142,123){\line(6,-5){120}}
              \put(118,102){\line(6,-5){95}}
              \put(94,82){\line(6,-5){71}}
              \put(70,62){\line(6,-5){47}}
              \put(47,42){\line(6,-5){23}}
              \put(123,116){\begin{picture}(40,14)
\thicklines       \put(0,12){\line(4,-1){40}}
                  \put(39,12){\line(-4,-1){38}}
                  \end{picture}}
              \put(99,95){\begin{picture}(40,14)
\thicklines       \put(0,12){\line(4,-1){40}}
                  \put(39,12){\line(-4,-1){38}}
                  \end{picture}}
              \put(51,55){\begin{picture}(40,14)
\thicklines       \put(0,12){\line(4,-1){40}}
                  \put(39,12){\line(-4,-1){38}}
                  \end{picture}}
              \put(26,35){\begin{picture}(40,14)
\thicklines       \put(0,12){\line(4,-1){40}}
                  \put(39,12){\line(-4,-1){38}}
                  \end{picture}}
              \put(75,75){\begin{picture}(40,14)
\thicklines       \put(0,12){\line(4,-1){40}}
                  \put(39,12){\line(-4,-1){38}}
                  \end{picture}}
\end{picture}}
\end{center}
\caption{An arbitrary right-branching TNCB structure} \label{bad_initial_guess}
\end{figure}
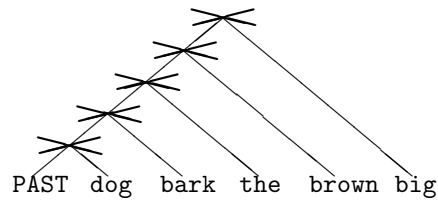

The generator cycles through two phases: a {\it test} phase and a {\it
rewrite} phase. Imagine a bag of signs, corresponding to ``the big brown dog
barked'', has been passed to the generation phase. The first step in the
generation process is to convert it into some arbitrary TNCB structure, say
the one in figure \ref{bad_initial_guess}. In order to verify whether this
structure is valid, we evaluate the TNCB. This is the test phase.  If the TNCB
evaluates successfully, the orthography of its value is the desired result. If
not, we enter the rewrite phase.

If we were continuing in the spirit of the original Shake-and-Bake generation
process, we would now form some arbitrary mutation of the TNCB and retest,
repeating this test-rewrite cycle until we either found a well-formed TNCB or
failed. However, this would also be intractable due to the undirectedness of
the search through the vast number of possibilities. Given the added
derivation information contained within TNCBs and the properties mentioned
above, we can direct this search by incrementally improving on previously
evaluated results.

We enter the rewrite phase, then, with an ill-formed TNCB. Each move operation
must improve it. Let us see why this is so.

The {\em move} operation maintains the same number of nodes in the tree. The
deletion of a maximal TNCB removes two ill-formed nodes (figure
\ref{deletion}). At the deletion site, a new undetermined node is created,
which
may or may not be ill-formed. At the destination site of the movement (whether
conjunction or adjunction), a new well-formed node is created.

The ancestors of the new well-formed node will be at least as well-formed
as they were prior to the movement. We can verify this by case:

\begin{enumerate}

\item When two maximal TNCBs are conjoined, nodes dominating the new node,
which were previously ill-formed, become undetermined. When re-evaluated, they
may remain ill-formed or some may now become well-formed.

\item When we adjoin a maximal TNCB within another TNCB, nodes dominating
the new well-formed node are disrupted. By dominance monotonicity, all nodes
which were disrupted by the adjunction must become well-formed
after re-evaluation. And nodes dominating the maximal disrupted node, which
were previously ill-formed, may become well-formed after re-evaluation.

\end{enumerate}
We thus see that rewriting and re-evaluating must improve the TNCB.

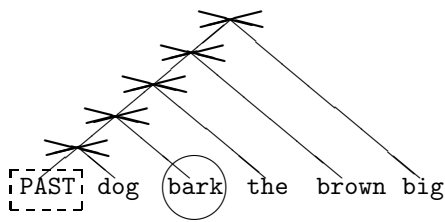
\begin{figure}[htbp]
\begin{center}
{\tt    \setlength{\unitlength}{0.60pt}
\begin{picture}(291,150)
\thinlines    \put(19,23){PAST}
              \put(68,23){dog}
              \put(112,23){bark}
              \put(162,23){the}
              \put(206,23){brown}
              \put(260,23){big}
              \put(31,32){\line(6,5){120}}
              \put(151,133){\line(6,-5){120}}
              \put(127,112){\line(6,-5){95}}
              \put(103,92){\line(6,-5){71}}
              \put(79,72){\line(6,-5){47}}
              \put(56,52){\line(6,-5){23}}
              \put(132,126){\begin{picture}(40,14)
\thicklines       \put(0,12){\line(4,-1){40}}
                  \put(39,12){\line(-4,-1){38}}
                  \end{picture}}
              \put(108,105){\begin{picture}(40,14)
\thicklines       \put(0,12){\line(4,-1){40}}
                  \put(39,12){\line(-4,-1){38}}
                  \end{picture}}
              \put(60,65){\begin{picture}(40,14)
\thicklines       \put(0,12){\line(4,-1){40}}
                  \put(39,12){\line(-4,-1){38}}
                  \end{picture}}
              \put(35,45){\begin{picture}(40,14)
\thicklines       \put(0,12){\line(4,-1){40}}
                  \put(39,12){\line(-4,-1){38}}
                  \end{picture}}
              \put(84,85){\begin{picture}(40,14)
\thicklines       \put(0,12){\line(4,-1){40}}
                  \put(39,12){\line(-4,-1){38}}
                  \end{picture}}
\thinlines    \put(13,13){\dashbox{5}(45,25){}}
              \put(129,27){\circle{38}}
\end{picture}}
\end{center}
\caption{The initial guess} \label{generation_step0}
\end{figure}

Let us further consider the contrived worst-case starting point
provided in figure
\ref{bad_initial_guess}. After the test phase, we discover that every single
interior node is ill-formed. We then scan the TNCB, say top-down from left to
right, looking for a maximal TNCB to move. In this case, the first move will
be {\it PAST} to {\it bark}, by conjunction (figure
\ref{generation_step0}).

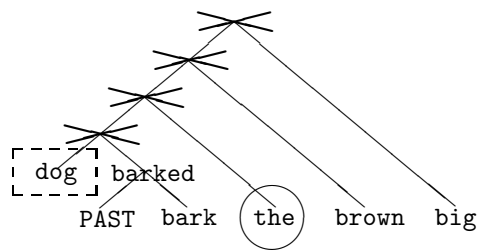
\begin{figure}[htbp]
\begin{center}
{\tt    \setlength{\unitlength}{0.70pt}
\begin{picture}(269,150)
\thinlines    \put(150,27){\circle{34}}
              \put(10,39){\dashbox{5}(45,25){}}
              \put(62,85){\begin{picture}(40,14)
\thicklines       \put(0,12){\line(4,-1){40}}
                  \put(39,12){\line(-4,-1){38}}
                  \end{picture}}
              \put(38,65){\begin{picture}(40,14)
\thicklines       \put(0,12){\line(4,-1){40}}
                  \put(39,12){\line(-4,-1){38}}
                  \end{picture}}
              \put(86,105){\begin{picture}(40,14)
\thicklines       \put(0,12){\line(4,-1){40}}
                  \put(39,12){\line(-4,-1){38}}
                  \end{picture}}
              \put(110,126){\begin{picture}(40,14)
\thicklines       \put(0,12){\line(4,-1){40}}
                  \put(39,12){\line(-4,-1){38}}
                  \end{picture}}
\thinlines    \put(57,72){\line(6,-5){47}}
              \put(81,92){\line(6,-5){71}}
              \put(105,112){\line(6,-5){95}}
              \put(129,133){\line(6,-5){120}}
              \put(238,23){big}
              \put(184,23){brown}
              \put(140,23){the}
              \put(90,23){bark}
              \put(22,48){dog}
              \put(46,22){PAST}
              \put(81,52){\line(-6,-5){24}}
              \put(63,47){barked}
              \put(130,133){\line(-6,-5){96}}
\end{picture}}
\end{center}
\caption{The TNCB after ``PAST'' is moved to ``bark''} \label{generation_step1}
\end{figure}

Once again, the test phase fails to provide a well-formed TNCB, so we repeat
the rewrite phase, this time finding {\it dog} to conjoin with {\it the}
(figure \ref{generation_step1} shows the state just after the second pass
through the test phase).

\begin{figure}[htbp]
\begin{center}
{\tt    \setlength{\unitlength}{0.70pt}
\begin{picture}(266,151)
\thinlines
              \put(10,24){PAST}
              \put(54,24){bark}
              \put(98,24){dog}
              \put(141,24){the}
              \put(175,24){brown}
              \put(230,24){big}
              \put(126,132){\line(1,-1){99}}
              \put(126,132){\line(-1,-1){99}}
              \put(105,113){\line(1,-1){78}}
              \put(88,93){\line(1,-1){58}}
              \put(48,53){\line(1,-1){20}}
              \put(126,53){\line(-1,-1){20}}
              \put(107,124){\begin{picture}(40,14)
\thicklines       \put(39,12){\line(-4,-1){38}}
                  \put(0,12){\line(4,-1){40}}
                  \end{picture}}
              \put(86,106){\begin{picture}(40,14)
\thicklines       \put(39,12){\line(-4,-1){38}}
                  \put(0,12){\line(4,-1){40}}
                  \end{picture}}
\thinlines    \put(170,15){\dashbox{5}(45,25){}}
              \put(110,27){\circle{34}}
              \put(26,55){barked}
              \put(107,55){the dog}
              \put(61,96){the dog}
              \put(66,85){barked}
\end{picture}}
\end{center}
\caption{The TNCB after ``dog'' is moved to ``the''} \label{generation_step2}
\end{figure}
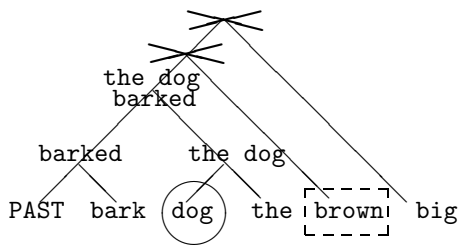

After further testing, we again re-enter the rewrite phase and this time note
that {\it brown} can be inserted in the maximal TNCB {\it the dog barked}
adjoined with {\it dog} (figure \ref{generation_step2}). Note how, after
combining {\it dog} and {\it the}, the parent sign reflects the correct
orthography even though they did not have the correct linear precedence.

After finding that {\it big} may not be conjoined with {\it the brown dog},
we try to adjoin it within the latter. Since it will combine with {\it brown
dog}, no adjunction to a lower TNCB is attempted.

\begin{figure}[htbp]
\begin{center}
{\tt    \setlength{\unitlength}{0.75pt}
\begin{picture}(278,146)
\thinlines
              \put(122,53){\oval(65,20)}
              \put(221,10){\dashbox{5}(32,20){}}
              \put(127,129){\line(-1,-1){101}}
              \put(127,129){\line(1,-1){101}}
	      \put(106,108){\line(1,-1){81}}
              \put(146,68){\line(-1,-1){40}}
              \put(127,49){\line(1,-1){22}}
              \put(47,49){\line(1,-1){22}}
              \put(107,122){\begin{picture}(40,14)
\thicklines       \put(39,12){\line(-4,-1){38}}
                  \put(0,12){\line(4,-1){40}}
                  \end{picture}}
              \put(228,19){big}
              \put(137,19){brown}
              \put(184,19){the}
              \put(57,19){bark}
              \put(100,19){dog}
              \put(10,19){PAST}
              \put(69,100){dog barked}
              \put(26,50){barked}
              \put(91,70){the brown dog}
              \put(91,50){brown dog}
              \put(72,110){the brown}
\end{picture}}
\end{center}
\caption{The TNCB after ``brown'' is moved to ``dog''} \label{generation_step3}
\end{figure}
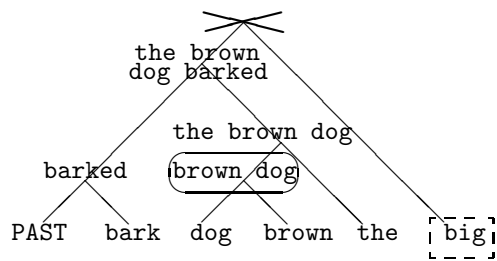

\begin{figure}[htbp]
\begin{center}
{\tt    \setlength{\unitlength}{0.75pt}
\begin{picture}(271,142)
\thinlines    \put(126,122){\line(-1,-1){101}}
              \put(126,122){\line(1,-1){101}}
              \put(165,82){\line(-1,-1){61}}
              \put(146,62){\line(1,-1){42}}
	      \put(126,42){\line(1,-1){22}}
              \put(47,42){\line(1,-1){22}}
              \put(57,13){bark}
              \put(10,13){PAST}
              \put(95,13){dog}
              \put(133,13){brown}
              \put(182,13){big}
              \put(225,13){the}
              \put(27,44){barked}
              \put(95,44){brown dog}
              \put(108,65){big brown dog}
              \put(108,85){the big brown dog}
              \put(50,124){the big brown dog barked}
\end{picture}}
\end{center}
\caption{The final TNCB after ``big'' is moved to ``brown dog''}
\label{generation_step4}
\end{figure}
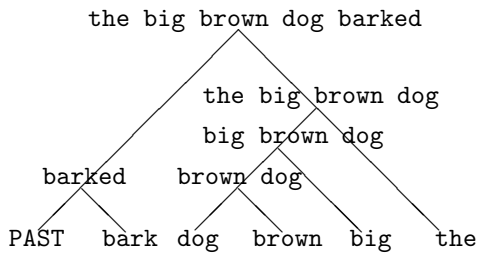

The final result is the TNCB in figure \ref{generation_step4}, whose
orthography is ``the big brown dog barked''.

We thus see that during generation, we formed a basic constituent, {\it the
dog}, and incrementally refined it by adjoining the modifiers in place. At the
heart of this approach is that, once well-formed, constituents can only grow;
they can never be dismantled.

Even if generation ultimately fails, maximal well-formed fragments will have
been built; the latter may be presented to the user, allowing graceful
degradation of output quality.

\section{Initialising the Generator}

\label{initialisation}

Considering the algorithm described above, we note that the number of rewrites
necessary to repair the initial guess is no more than the number of ill-formed
TNCBs. This can never exceed the number of interior nodes of the TNCB formed
from $n$ lexical signs (i.e. $n-2$). Consequently,
the better formed the initial TNCB used by the generator, the fewer the number
of rewrites required to complete generation. In the last section, we
deliberately illustrated an initial guess which was as bad as possible. In
this section, we consider a heuristic for producing a motivated guess for the
initial TNCB.

Consider the TNCBs in figure \ref{equivalences}. If we interpret the S, O and
V as Subject, Object and Verb we can observe an equivalence between the
structures with the bracketings: (S (V O)), (S (O V)), ((V O) S), and ((O V)
S). The implication of this equivalence is that if, say, we are translating
into a (S (V O)) language from a head-final language and have isomorphic
dominance structures between the source and target parses, then simply
mirroring the source parse structure in the initial target TNCB will provide a
correct initial guess. For example, the English sentence (5): \\
\\
(5) the book is red\\
\\
has a corresponding Japanese equivalent (6): \\
\\
\begin{tabular}{ll@{ }l@{ }l@{ }l}
(6) & ((hon & wa) & (akai & desu)) \\
    & ((book & TOP)  & (red  & is))
\end{tabular}\\

If we mirror the Japanese bracketing structure in English to form the initial
TNCB, we obtain: ((book the) (red is)). This will produce the correct answer
in the test phase of generation without the need to rewrite at all.

Even if there is not an exact isomorphism between the source and target
commutative bracketings, the first guess is still reasonable as long as the
majority of child commutative bracketings in the target language are
isomorphic with their equivalents in the source language. Consider the French
sentence: \\
\\
\begin{tabular}{ll@{ }l@{ }l@{ }l@{ }l}
(7) & ((le  & ((grand & chien) & brun)) & aboya) \\
(8) & ((the & ((big & dog) & brown)) & barked)
\end{tabular}\\

The TNCB implied by the bracketing in (8) is equivalent to that in figure
\ref{generation_step3} and requires just one rewrite in order to make it
well-formed. We thus see how the TNCBs can mirror the dominance information in
the source language parse in order to furnish the generator with a good
initial guess. On the other hand, no matter how the SL and TL structures
differ, the algorithm will still operate correctly with polynomial complexity.
Structural transfer can be incorporated to improve the efficiency of
generation, but it is never necessary for correctness or even tractability.

\section{The Complexity of the Generator}

The theoretical complexity of the generator is $O(n^4)$, where $n$ is the size
of the input. We give an informal argument for this. The complexity of the
test phase is the number of evaluations that have to be made. Each node must
be tested no more than twice in the worst case (due to precedence
monotonicity), as one might have to try to combine its children in either
direction according to the grammar rules.  There are always exactly $n-1$
non-leaf nodes, so the complexity of the test phase is $O(n)$. The complexity
of the rewrite phase is that of locating the two TNCBs to be combined. In the
worst case, we can imagine picking an arbitrary child TNCB ($O(n)$) and then
trying to find another one with which it combines ($O(n)$). The complexity of
this phase is therefore the product of the picking and combining complexities,
i.e. $O(n^2)$. The combined complexity of the test-rewrite cycle is thus
$O(n^3)$.  Now, in section \ref{initialisation}, we argued that no more than
$n-1$ rewrites would ever be necessary, thus the overall complexity of
generation (even when no solution is found) is $O(n^4)$.

Average case complexity is dependent on the quality of the first guess, how
rapidly the TNCB structure is actually improved, and to what extent the TNCB
must be re-evaluated after rewriting. In the SLEMaT system
\cite{Poznanski:SLEMaT}, we have tried to form a good initial guess by
mirroring the source structure in the target TNCB, and allowing some local
structural modifications in the bilingual equivalences.

Structural transfer operations only affect the efficiency and not the
functionality of generation.  Transfer specifications may be
incrementally refined and empirically tested for efficiency.  Since
complete specification of transfer operations is not required for
correct generation of grammatical target text, the version of
Shake-and-Bake translation presented here maintains its advantage over
traditional transfer models, in this respect.

The monotonicity constraints, on the other hand, might constitute a dilution
of the Shake-and-Bake ideal of independent grammars. For instance, precedence
monotonicity requires that the status of a clause (strictly, its
lexical head) as main or subordinate has to be transferred into German.
It is not that the transfer of information {\it per se} compromises the
ideal --- such information must often appear in transfer entries to avoid
grammatical but incorrect translation (e.g. {\it a great man} translated as
{\it un homme grand}). The problem is justifying the main/subordinate
distinction in every language that we might wish to translate into German.
This distinction can be justified monolingually for the other languages
that we treat (English, French, and Japanese). Whether the constraints
will ultimately require monolingual grammars to be enriched with
entirely unmotivated features will only become clear as translation coverage
is extended and new language pairs are added.

\section{Conclusion}

We have presented a polynomial complexity generation algorithm which can form
part of any Shake-and-Bake style MT system with suitable grammars and
information transfer. The transfer module is free to attempt
structural transfer in order to produce the best possible first guess. We
tested a TNCB-based generator in the SLEMaT MT system
with the pathological cases described in \cite{Brew:Cat}
against Whitelock's original generation algorithm, and have
obtained speed improvements of several orders of magnitude. Somewhat more
surprisingly, even for short sentences which were not problematic for
Whitelock's system, the generation component has performed consistently
better.

\bibliographystyle{acl}

\end{document}